\pdfoutput=1
\documentclass[11pt,a4paper]{article}
\usepackage[hyperref]{emnlp2020}
\usepackage{times}
\usepackage{graphicx}
\usepackage{latexsym}
\usepackage{natbib}
\usepackage{amsmath}
\usepackage{amssymb}
\usepackage{mathtools}
\usepackage{anyfontsize}
\usepackage{booktabs}
\usepackage{csquotes}
\usepackage{multirow}
\usepackage{array}
\usepackage{parskip}
\usepackage{setspace}

\makeatletter
\renewenvironment*{displayquote}
  {\begingroup \setlength{\leftmargini}{0.43cm}\csq@getcargs{\csq@bdquote{}{}}}
  {\csq@edquote \endgroup}
\makeatother
\usepackage{microtype}

\aclfinalcopy 
\def\aclpaperid{3292} 

\usepackage{amsmath,amsfonts,bm}

\def\eqref#1{equation~\ref{#1}}

\def\1{\bm{1}}

\def\rd{{\textnormal{d}}}

\def\rw{{\textnormal{w}}}

\def\rz{{\textnormal{z}}}

\def\rvd{{\mathbf{d}}}

\def\rvw{{\mathbf{w}}}

\def\rvz{{\mathbf{z}}}

\def\rmI{{\mathbf{I}}}

\def\vmu{{\bm{\mu}}}
\def\vsigma{{\bm{\sigma}}}

\def\vb{{\bm{b}}}
\def\vc{{\bm{c}}}

\def\ve{{\bm{e}}}
\def\vf{{\bm{f}}}

\def\vh{{\bm{h}}}
\def\vi{{\bm{i}}}

\def\vo{{\bm{o}}}

\def\vr{{\bm{r}}}
\def\vs{{\bm{s}}}

\def\vz{{\bm{z}}}

\def\mA{{\bm{A}}}

\def\mI{{\bm{I}}}

\def\mU{{\bm{U}}}
\def\mV{{\bm{V}}}
\def\mW{{\bm{W}}}

\DeclareMathAlphabet{\mathsfit}{\encodingdefault}{\sfdefault}{m}{sl}
\SetMathAlphabet{\mathsfit}{bold}{\encodingdefault}{\sfdefault}{bx}{n}

\def\sR{{\mathbb{R}}}

\newcommand*\rot{\rotatebox[origin=c]{90}}
\newcolumntype{C}[1]{>{\centering\let\newline\\\arraybackslash\hspace{0pt}}m{#1}}
\newcolumntype{L}[1]{>{\raggedright\let\newline\\\arraybackslash\hspace{0pt}}m{#1}}

\title{VCDM: Leveraging Variational Bi-encoding and Deep Contextualized Word Representations for Improved Definition Modeling}

\author{Machel Reid \qquad Edison Marrese-Taylor \qquad Yutaka Matsuo\\
Graduate School of Engineering\\
The University of Tokyo\\
{\tt machelreid2004@gmail.com}\\ {\tt\{emarrese, matsuo\}@weblab.t.u-tokyo.ac.jp} \\}

\date{}

\begin{document}
\maketitle
\begin{abstract}
In this paper we tackle the task of definition modeling, where the goal is to learn to generate definitions of words and phrases. Existing approaches for this task are discriminative, combining distributional and lexical semantics in an implicit rather than direct way. To tackle this issue we propose a generative model for the task, introducing a continuous latent variable to explicitly model the underlying relationship between a phrase used within a context and its definition. We rely on variational inference for estimation and leverage contextualized word embeddings for improved performance. Our approach is evaluated on four existing challenging benchmarks with the addition of two new datasets, \textsc{Cambridge} and the first non-English corpus \textsc{Robert}, which we release to complement our empirical study. Our Variational Contextual Definition Modeler (VCDM) achieves state-of-the-art performance in terms of automatic and human evaluation metrics, demonstrating the effectiveness of our approach.\footnote{We release the code at: \url{https://github.com/machelreid/vcdm}}
\end{abstract}

\section{Introduction}
In most current NLP tasks, fixed-length vector representations of words, word embeddings, are used to represent some form of the meaning of the word. In the case of humans, however, oftentimes we will use a sequence of words known as a \textit{definition} ---a statement of the meaning for a term--- to express meanings of terms (words, phrases, or symbols). It is with this in mind that the question of ``Can machines define?'' is aimed to be answered with the task of \textbf{definition modeling} \citep{noraset}.

Definition modeling can be framed as a task of conditional generation, in which the definition $\rvd$ of the word or phrase is generated given a conditioning variable $\rvw$ such as a word's associated word embedding or other representations of context. Current approaches for this task \citep{noraset, gadetsky, niLearningExplainNonStandard2017a, ishiwatari-etal-2019-learning} are mainly encoder-decoder based, in which one encodes a contextual representation for a word/phrase $\rvw$ using a variety of features such as context or character composition, and uses the contextual representation(s) to generate the definition $\rvd$. 

Despite the relative success of existing approaches for definition modelling, their discriminative nature ---where distributional-derived information is at one end of the model and lexical information is at the other--- limits their power as the underlying semantic representations of the distributional and lexical information are learned in an implicit rather than direct way. For example, although \citet{ishiwatari-etal-2019-learning} successfully showed that both local and global contexts are useful to disambiguate meanings of phrases in certain cases, their approach heavily relies on an attention mechanism to identify semantic alignments between the input phrase and the output definition, which may introduce noise and ultimately be insufficient to capture the entire meaning of each phrase-definition pair.

To tackle this issue, we propose to explicitly model the underlying semantics of  phrase-definition pairs by introducing a continuous latent variable $\rvz$ over a definition space, which is used in conjunction with $\rvw$ to guide the generation of definition $\rvd$. The introduction of this latent representation enables us to treat it as a global defining signal during the generation process, complementing existing alignment mechanisms such as the attention. 


Although the latent definition variable enables us to explicitly model underlying semantics of context-definition pairs, the incorporation of it into the task renders the posterior intractable. In this paper we recur to variational inference to estimate this intractable posterior, effectively making our model a Conditional Variational Autoencoder and evolving the generation process from $p(\rvd|\rvw)$ to $p(\rvd|\rvw, \rvz)$.



We also note that existing approaches for definition modelling heavily rely on word embeddings, which due to their fixed nature can only capture so much of the semantics, being known to offer limited capabilities when dealing with polysemy. Considering the success of pretrained deep contextualized word representations which by specifically addressing these limitations have been shown to improve performance on a variety of downstream NLP tasks \citep{peters-etal-2018-deep, devlin2018bert}, in this paper we propose a mechanism to integrate deep contextualized word representations in the definition modelling task. Specifically, we successfully leverage BERT \citep{devlin2018bert} as our contextual encoder and our definition encoder to produce representations for $\rvw$ and $\rvd$ respectively. 


Finally, we develop two new datasets for this task, one derived from the Cambridge Dictionary
\footnote{\url{https://dictionary.cambridge.org/}}, and the other derived from Le Petit Robert\footnote{\url{https://dictionnaire.lerobert.com/}}. In summary, our contributions are:
\begin{itemize}
	\item \textbf{Model:} We propose a novel approach for the task of definition modeling, leveraging deep contextualized word representation and the variational encoder-decoder architecture. We achieve new state-of-the-art performance on the definition modeling task, outperforming the previous state-of-the-art by as much as 9 BLEU points on the \textsc{Oxford} dataset and 22 BLEU points on the \textsc{Robert} dataset.  

	\item \textbf{Datasets:} We develop two new datasets \textsc{Cambridge} and \textsc{Robert} for this task. With \textsc{Robert}, a French dataset, being the first non-English dataset developed for this task.
\end{itemize}
Datasets and pre-trained models will be publicly released to the greater NLP community to help facilitate further advances on this task upon acceptance of this paper.
\section{Related Work}

Our work is related to the seminal paper by \citet{hill-dictionary-2016}, who proposed using the definitions found in everyday dictionaries as a means of bridging existing gaps between lexical and phrasal semantics. Effectively, they train a language model to map dictionary definitions to lexical representations of words, presenting the task of reverse dictionaries, where the goal is to return the name of a concept given a definition.


\citet{noraset} later introduced the task of definition modeling, in which a model is tasked with generating a definition for a given word, given its respective embedding. The authors argued that, compared to other related tasks such as word similarity or analogical relatedness, definition generation can be considered a more transparent view of the information captured by an embedding. However, this method does not incorporate contextual information, preventing it from generating appropriate definitions for polysemic words. Addresing this, \citet{gadetsky} studied the problem of polysemy in definition modeling, introducing an attention-based model which uses contextual information determine components in the embedding which may refer to a relevant word meaning. 



\citet{niLearningExplainNonStandard2017a} explore a different but related problem, proposing an approach for automatically explaining non-standard English expressions (i.e. slang) in a given sentence. They present a hybrid word-character sequence-to-sequence model that directly explains unseen non-standard expression, garnering reasonable definitions of expressions given their context.


More recently, \citet{ishiwatari-etal-2019-learning} have tackled some of the limitations of previous works on definition modelling and non-standard English expression explanation. Concretely, they note that whenever it is not possible to figure out the meaning of a given expression from its immediate local context, it is common to consult dictionaries for definitions or search documents or the web to find other global context to help in interpretation. In light of this, they introduce the task of describing a given phrase in natural language, based on its local and global contexts. To tackle this the authors introduce a model which consists of two context encoders (one for the local context, and one for the global context) as well as a description decoder. Our proposed model, uses a more practical variational encoder-decoder framework, allowing us to take advantage of explicitly modeling the phrase-definition relationship, while also leveraging deep contextualized word representations for more informative context representations.


Finally, our model is also related to \citet{sohnLearningStructuredOutput2015}, in which Conditional Variational Autoencoders (CVAEs) ---an extension of the original variational autoencoder (VAE) \citep{kingmaAutoEncodingVariationalBayes2014}--- were proposed for generating diverse structured output, mainly in the context of image generation, and visual object segmentation and labeling. Our work is also related to CVAE models that have been developed for the domain of natural language processing, specifically \citet{zhang-etal-2016-variational-neural} who proposed a CVAE in the context of Neural Machine Translation (NMT). As the usage of VAEs has become relatively common, we will omit a detailed explanation of these models, referring readers to \citet{kingmaAutoEncodingVariationalBayes2014}.

\section{Proposed Approach}

In a way analogous to previous work \citep{noraset}, our proposed approach is a generative probabilistic model for word definitions, in which the goal is to estimate the probability of generating a definition $d$, given an input $w$. Concretely, we propose to directly capture the joint semantics of the $(w,d)$ pairs by introducing a latent variable $\rvz$ to model the underlying \textit{definition space}. Our proposed generative process can be formulated as follows:
\begin{align}
        p(\rd|\rw) &= \int_z p(\rd, z|\rw)d_z  = \int_z p(\rd|z,\rw)p(z|\rw)d_z
\end{align}
where the conditional probability $p(\rd|\rw)$ evolves into $p(\rd|\rw, \rz)$, and the generation of the definition $\rd$ is now conditioned on both the input variable $\rw$ and our introduced continuous latent variable $\rz$. 


Since the introduction of our latent variable makes posterior inference intractable, in this paper we resort to variational inference to perform posterior approximation. Effectively, this makes our proposed generative model a CVAE \citep{sohnLearningStructuredOutput2015} such that the variational lower bound can be formulated as follows: 
\begin{multline}
	\log p(\rd|\rw) \geq \mathbb{E}_{\rz\sim q(\rz)}\big[\log p(\rd|\rw,\rz)\big] \\- D_{\text{KL}}\big[q(\rz) || p(\rz|\rw)\big]
\end{multline}
where $q(\rz)$ is the introduced variational approximation to the intractable posterior $p(\rz|\rw,\rd)$ and $p(\rz|\rw)$ is the prior distribution. Following previous work \citep{sohnLearningStructuredOutput2015,zhang-etal-2016-variational-neural} we let $\rvz$, $\rvw$ and $\rvd$ be random vectors associated to $\rz$, $\rw$ and $\rd$ respectively, and utilize neural networks to estimate the following components.
\begin{itemize}
    \item $q(\rz) \approx q_\phi(\rvz|\rvw,\rvd)$ is our variational approximation for the intractable posterior (the \textit{recognition network}), which we model with a neural network with parameters $\phi$. This makes $q_\phi(\rvz|\rvw,\rvd)$ a \textit{neural definition inferer}.
    \item $p(\rz|\rw) \approx p_\theta(\rvz|\rvw)$ is a \textit{(conditional) prior network}, parameterized by $\theta$, which in our case can be regarded as a \textit{neural definition prior}.
    \item $p(\rd|\rw,\rz) \approx p_\theta(\rvd|\rvw,\rvz)$ is a \textit{generation network}, parameterized by $\theta$ which acts as a \textit{variational definition modeler}. 
\end{itemize}
In the following subsections we give details on how we specifically model each one of these components.
With this in mind, we develop the following architecture comprised of 3 major components:
\begin{itemize}
	\item \textbf{Encoders} - This component is comprised of two encoders - one \textit{context encoder} to produce a representation for $\rvw$ and another \textit{definition encoder} to produce a representation for $\rvd$ (Section \ref{encoders}).
	\item \textbf{Neural Definition Inferer} - This component \textit{infers} the latent representation $\rvz$ from the representation of the word/phrase ---the explicitly modeled prior $p_\theta(\rvz|\rvw)$--- and in conjunction with the definition (the approximated posterior $q_\phi(\rvz|\rvd, \rvw)$ )  (Section \ref{inferer}).
	\item \textbf{Variational Definiton Modeler} - This component can be viewed as a decoder which takes in latent representation $\rvz$ to guide the generation of the target sentence, essentially $p_\theta(\rvd|\rvw,\rvz)$ (Section \ref{vdm_section}).
\end{itemize}

\paragraph{Notation} For clarity when explaining the approach, we define the notation conventions we will follow, namely: $d$ refers to dimensions, $\vc$ refers to the context vectors produced by the attention mechanism, $g$ refers to projections/activations, and $\vh$ refers to sets of vectors.

\subsection{Encoders} \label{encoders}
\subsubsection{Context Encoder} \label{sec:pcp}

To encode the sequence in which the word in question is used, we adopt the BERT \citep{devlin2018bert} architecture, which is comprised of multiple Transformer \citep{vaswani2017attention} encoder layers pretrained on a masked-language modeling task to encode deep contextual word representations for a given sequence. BERT has also shown to be able to model the relationship between two tasks on pair-wise natural language understanding tasks. It is to this end that we propose the construction of \textit{phrase-context pairs} to leverage this property in the context encoding process. 


Inspired by \textit{context-gloss pairs} \citep{Huang_2019} for the task of Word-Sense Disambiguation (WSD), we construct \textit{phrase-context pairs} for our task of definition modeling. Often, there are differences in the word or phrase that we aim to define, and the lexeme form that is used in the context sentence. For example, the lemma \textit{run} has the following forms: \textit{run}, \textit{runs}, \textit{ran} and \textit{running}, which all represent the same lexeme. To account for these discrepancies between the lemma and the lexeme form, we construct the aforementioned \textit{phrase-context pairs}, which are constructed by simply inserting a separator token, denoted as [SEP], between the word/phrase and the context sentence. Below we show how this process would work for an example taken from Cambridge dictionary dataset for the lemma \textit{leave}:
\begin{displayquote}
    \fontsize{9.75}{12}
    He \textit{left} a wife and two children.\\
    $\hookrightarrow$ \textit{leave} [SEP] He \textit{left} a wife and two children.
\end{displayquote}

This form of construction for the \textit{phrase-context pairs} comes with the added benefit of querying a sentence for a definition by simply prepending the word/phrase and a seperator token to the context sequence. As we use BERT as our encoder, we are able to leverage its self attentive nature to produce a representation of the word or phrase in question with respect to the context sentence. 

As we initialize our context encoder with BERT, the phrase-context pair sequence $c~=~[w_t, \text{[SEP]}, c_2, \dots c_{M_c}]$, containing word or phrase $w_t$, is prepended by a [CLS] token and is appended by a [SEP] token, making $c_0 = \text{[CLS]}$ and $c_{M_c} = \text{[SEP]}$.

We define this \textbf{context encoder} as $T_c$, which takes in the context sequence $c$, and returns a sequence of annotation vectors for each token in $c$. We denote these annotation vectors as as $\vh_c$, where $\{\vh_c^{(i)}\}_{i=0}^{M_c} \in \mathbb{R}^{d_c}$ and
\begin{equation}
    \vr_{w_t} = T_c(c)[t]
\end{equation}
is the $t^\text{th}$ representation in $h_c$, representing $w_t$. In the case that $w_t$ is split into multiple subword tokens by the BERT tokenizer, we set the word representation to be the mean of each of its subword representations. Namely, in the case that $w_t$ is comprised of the $n^\text{th}$ to the $m^\text{th}$ subtokens,
\begin{equation}
    \vr_{w_t} = \frac{1}{m-n}\textstyle\sum^{m}_{i=n}T_c(c)[i]
\end{equation}

\subsubsection{Definition Encoder}

The definition encoder, which we denote as $T_d$, is also initialized with BERT. This encoder takes in the definition sequence $d = [d_0, d_1, \dots, d_{M_e}]$ as input and represents $d$ as:
\begin{equation}
    \vr_d = T_d(d)[0]
\end{equation}
where $\vr_d \in \mathbb{R}^{d_e}$. We take the representation (corresponding to the preprended [CLS] token) as a representation for the entire definition sequence.

\subsection{Neural Definition Inferer}\label{inferer}
\label{sec:inferer}

We formulate the posterior distribution $q_\phi(\rvz|\rvd,\rvw)$ and prior distribution $p_\theta(\rvz|\rvw)$ as multivariate Gaussians with a diagonal covariance matrices. To model these distributions we make use of neural networks, following \citet{zhang-etal-2016-variational-neural}.

\subsubsection{Neural Definition Posterior}

As modeling the true posterior $p(\rvz|\rvd,\rvw)$ is generally intractable, to approximate this true posterior, we use a variational distribution, formulated as the following multivariate Gaussian:
\begin{equation}
	q_\phi(\rvz|\rvd,\rvw) = \mathcal{N}(\rvz;\mu(\rvd,\rvw),\sigma(\rvd,\rvw)^2\rmI)
\end{equation}
which is parameterized by the mean and $\mu(\rvd,\rvw)$ standard deviation $\sigma(\rvd,\rvw)$, both which are treated as functions of definition $\rvd$ and phrase $\rvw$ parameterized by neural networks.

From the neural encoding mechanisms, we gather the definition representation $\vr_d$, and the context representation $\vr_{w_t}$. We then concatenate $\vr_d$ and $\vr_{w_t}$ and  project the resulting vector onto our latent space, setting $\vh_z~=~g(\mW_z[\vr_{w_t};\vr_d]+\vb_z)$, where $\mW_z \in \sR^{d_z \times (d _e+d_c)}$ is a trainable weight matrix, $\vb_z \in \sR^{d_z}$ is a trainable bias vector and $g(\cdot)$ represents a non-linearity activation. In our experiments we set $g(\cdot)$ to be the \textit{tanh}$(\cdot)$ activation function, following previous work. 

To attain the aforementioned mean and variance vectors parameterizing the variational distribution, setting $\vmu = \mW_\mu \vh_z + \vb_\mu$ and $\log \vsigma^2 = \mW_\sigma \vh_z + \vb_\sigma$, where $\mW_\mu, \mW_\sigma \in \mathbb{R}^{d_z \times d_z}$ are trainable weight matrices parameterizing the projection and $\vb_\mu,\vb_\sigma \in \mathbb{R}^{d_z}$ are bias vectors.

In order to make the parameters $\theta$ differentiable for gradient descent optimization, we use the ``reparameterization trick" \citep{kingmaAutoEncodingVariationalBayes2014} setting $\vz = \vmu + \vsigma \cdot \ve$, where $\ve \sim \mathcal{N}(0,\mI)$ is a noise variable sampled from a multivariate Gaussian distribution to derive our latent vector $\vz$.

\subsubsection{Neural Definition Prior}
Our prior is a conditional distribution formulated as the following multivariate Gaussian:
\begin{equation}
	p_\theta(\rvz|\rvw) = \mathcal{N}(\rvz ;\mu^{\prime}(\rvw),\sigma^{\prime}(\rvw)^2\rmI)
\end{equation}
which is parameterized by $\mu^{\prime}(\cdot)$, and $\sigma^{\prime}(\cdot)$ which are both solely functions of phrase $\rvw$. In a similar fashion to the \textit{neural definition posterior}, we make use of a linear projection to project $\vr_{w_t}$ to the mean vector $\vmu^{\prime}$ and another linear projection to derive the log variance vector. During inference (at test time) when sampling from $p_\theta(\rvz|\rvw)$, we set our latent vector $\vz$ to be the mean vector $\vmu^{\prime}$.

To initialize the decoding procedure detailed in the next subsection, we feed the latent representation $\vz$ and project it to the decoding space setting $\vh_d^{\prime}~=~g(\mW_d\vz + \vb_d)$, where $\mW_d \in \mathbb{R}^{d_d \times d_z}$ and $\vb_d \in \mathbb{R}^{d_d}$.

%

\subsection{Variational Definition Modeler} \label{vdm_section} 

Given phrase $\rvw$ and latent representation $\rvz$, the process of definition modeling can be formulated as the following conditional language model:
\begin{align}
    p(\rvd|\rvw,\rvz) &= \textstyle\prod_{j=1}^{M_d} p(d_j | d_{<j},\rvz,\rvw) \\
    p(d_j | d_{<j},\rvz,\rvw) &= g_d(\vs_j,\vc_j)
\end{align}
where $g_d$ is a feed-forward neural network which returns a distribution over the elements in the decoder vocabulary given the context vector $\vc_j$ (see Eq. \ref{eq:att}) and decoder state $\vs_j$.

During generation of the definition sequence, we want the decoder to rely on all of the encoded components at each timestep. We modify the LSTM Cell \citep{hochreiter1997long} to encompass previous context vector $\vc_{j-1}$, and the projected latent definition representation $\vh_d^{\prime}$.

Intuitively, at each timestep $j$, we want the generated token to have the ability to rely on each of these components in the case that the previous hidden state and/or generated token does not provide enough information or misleads the accurate generation of the next token. We refer to this modified cell as the Variational Contextual Definition Modeler (VCDM) Cell, and the resulting decoder as a VCDM-RNN. The VCDM Cell calculates the decoder hidden state $\vs_j$ as follows\footnote{Note: For clarity, we omit the bias terms in Equations \ref{eq:vdmcell}-\ref{eq:vdmcell.end}}:
{\fontsize{9.75}{12}
\begin{align} \label{eq:vdmcell} 
    \vi_j &= \sigma(\mW E_{d_j} + \mU \vs_{j-1} + \mA c_{j-1}+ \mV \vh'_d)\\
    \vf_j &= \sigma(\mW_f E_{d_j} + \mU_f \vs_{j-1} + \mA_f \vc_{j-1}+ \mV_f \vh'_d)\\
    \vo_j &= \sigma(\mW_o E_{d_j} + \mU_o \vs_{j-1} + \mA_o \vc_{j-1}+ \mV_o \vh'_d)\\
    \bm{\tilde{C}}_j &= g(\mW_g E_{d_j} + \mU_g \vs_{j-1} + \mA_g \vc_{j-1}+ \mV_g \vh'_d)\\
    \bm{C}_j &= \sigma(\vf_j \cdot \bm{C}_{j-1} + \vi_j \cdot \bm{\tilde{C}}_j)\\ \label{eq:vdmcell.end}
    \vs_j &= g(\bm{C}_j) \cdot \vo_j
\end{align}}
where $E_{d_j} \in \mathbb{R}^{d_w} $ is the embedding for the target word, $\mW$,$\mW_f$,$\mW_o$,$\mW_g \in \sR^{d_d \times d_w}$, $\mU$,$\mU_f$,$\mU_o$,$\mU_g \in \sR^{d_d \times d_d}$,  $\mA$,$\mA_f$,$\mA_o$,$\mA_g \in \sR^{d_d \times d_d}$, and $\mV$,$\mV_f$,$\mV_o$,$\mV_g \in \sR^{d_d \times d_d}$ are trainable weight matrices parameterizing the RNN cell.

Additionally, at each decoder timestep $j$ we attend to the set of annotation vectors $\vh_c$ produced by the last layer of the context encoder. To compute context vector $\vc_j$, we use general attention \citep{luong2015effective} shown below:
\begin{align} \label{eq:att}
    \vc_j & = \textstyle \sum^T_{i=1}\alpha_i \vh_c^{(i)} \\
    \alpha_i & = \text{softmax}(\vs_j^{\top} \mW_a \vh_c^{(i)})
\end{align}
where $\mW_a \in \mathbb{R}^{d_d \times d_c}$, and $\alpha_i$ can be viewed as an alignment over $\vh_c$ and $\vc_j$ as a vector capturing the encoder hidden states scaled by this alignment.

\subsection{Optimization challenges}


Despite the VAE's appeal as a tool to learn unsupervised representations through the use of latent variables, these models are often found to ignore latent variables when using powerful generators. To overcome this issue of ``posterior collapse"   \citep{bowmanGeneratingSentencesContinuous2016}, we incorporate the following heuristics: (1) annealing the KL term from 0 to 1 using a sigmoid annealing schedule, following \citet{bowmanGeneratingSentencesContinuous2016} and (2) thresholding the KL term in the objective function with a constant $\lambda$ using  the ``free bits" technique \citep{kingmaImprovedVariationalInference2016}. With these changes, our objective function is modified to become the following:
{\fontsize{9.85}{12}
    \begin{multline}
        \mathcal{L}(\theta,\phi) = - \mathbb{E}_{\rvz\sim q_\phi(\rvz|\rvw,\rvd)}\big[\log p_\theta(\rvd|\rvw,\rvz)\big] \\ + \gamma \sum _i\mathrm{max}(\lambda,D_{\mathrm{KL}}(q_\phi(\rvz_i|\rvw,\rvd) || p(\rvz_i|\rvw)))
    \end{multline}
}
Where $\gamma$ denotes the annealing term that follows the sigmoid schedule and $\lambda$ denotes the \textit{target rate}, and the sub-index $i$ denotes the $i^{th}$ dimension of the latent vector $\rvz$. In our experiments, we set the total $\lambda = 1$.


\section{Empirical Study}

\subsection{Data}

To evaluate our approach we make use of the following previously released datasets: \textsc{Oxford} \citep{gadetsky} built from Oxford Dictionaries\footnote{\url{oxforddictionaries.com}}, \textsc{Urban} built from the Urban Dictionary\footnote{\url{urbandictionary.com}}, and \textsc{Wikipedia} \citep{ishiwatari-etal-2019-learning} built from Wikipedia.

The task of definition modeling with respect to each of the aforementioned datasets can be regarded as three separate domains, in which (1) \textsc{Oxford} can be viewed as a corpus of ``traditional" dictionary definitions, where most common words in a given language are contained, (2) \textsc{Urban} can be viewed as a corpus of ``uncommon  , slang words in which one often has to use context and subword information to decipher the meaning, and (3) \textsc{Wikipedia} can be viewed almost as a description generation task of named entities, conditioned on the given context.

\begin{table} 
	\centering
	\tabcolsep=0.13cm
    \scalebox{0.65}{
    \begin{tabular}{ccc|ccc}
        \toprule[0.15em]
        {\large\textsc{Cambridge}} & \multicolumn{2}{c}{\textbf{Count}} & \multicolumn{3}{c}{\textbf{Length}} \\
        \cmidrule{2-6}
        \textbf{Partition} & \textbf{Phrases}     & \textbf{Examples}     & \textbf{Phrase} & \textbf{Context} & \textbf{Definition} \\
        \midrule
        Train & 21,993      & 42,689       & 1.01            & 9.14 $\pm$ 4.27  & 11.64 $\pm$ 6.75    \\
        Valid & 4,671       & 5,335        & 1.00            & 9.25 $\pm$ 4.24  & 11.69 $\pm$ 6.80    \\
        Test & 4,670       & 5,337        & 1.00            & 9.20 $\pm$ 4.32  & 11.60 $\pm$ 6.75 \\
        \midrule
        Overall & 24,557      & 53,361       & 1.01            & 9.16 $\pm$ 4.27  & 11.64 $\pm$ 6.76    \\
        \bottomrule
        \toprule
        {\large\textsc{Robert}} & \multicolumn{2}{c}{\textbf{Count}} & \multicolumn{3}{c}{\textbf{Length}} \\
        \cmidrule{2-6}
        \textbf{Partition} & \textbf{Phrases}     & \textbf{Examples}     & \textbf{Phrase} & \textbf{Context} & \textbf{Definition} \\
        \midrule
        Train & 30,049      & 71,073       & 1.00            & 10.51 $\pm$ 7.36 & 7.97 $\pm$ 4.95    \\
        Valid & 6,992       & 8,884        & 1.00            & 10.55 $\pm$ 7.47  & 7.93 $\pm$ 4.90    \\
        Test & 6,985       & 8,884        & 1.00            & 10.46 $\pm$ 7.36  & 8.03 $\pm$ 5.00 \\
        \midrule
        Overall & 33,507     & 88,842      & 1.00            & 10.51 $\pm$ 7.38  & 7.97 $\pm$ 4.94    \\
        \bottomrule[0.15em]
        \end{tabular}
    }
    \caption{Statistics for \textsc{Cambridge} and \textsc{Robert}. The number of individual phrases, number of examples, and the mean and s.d. of the lengths of each partition of the dataset are reported}\label{tb:data}
\end{table} 

In addition to these datasets, we also develop the \textsc{Cambridge} (English) and \textsc{Robert} (French) dataset. We collect this data from the online version of the Cambridge Dictionary\footnote{\url{dictionary.cambridge.org}} and Le Petit Robert\footnote{\url{https://dictionnaire.lerobert.com/}}. Following the spirit of previously released datasets, we include three components for each example: (1) the word or phrase being defined, (2) an example (context sentence) in which it is contained and (3) its corresponding definition. These datasets can be seen as an addition to the domain of ``traditional'' dictionary definitions, with \textsc{Robert} being the first non-English dataset. Please refer to Table~\ref{tb:data} for statistics regarding these datasets. 

\subsection{Experiments}
\subsubsection{Our Model: VCDM}

We initialize each of our encoders with BERT-base-uncased (or in the case of \textsc{Robert}, CamemBERT-base \citep{martin2019camembert}), setting $d_e, d_c = 768$. We set latent dimension $d_z = 83$, and the LSTM decoder's hidden size $d_d = 512$ with an output vocabulary size of 10k, initializing embeddings with Word2Vec \citep{mikolov2013efficient}. We perform gradient descent using the Adam optimizer \citep{kingma2014adam} with its default hyperparameters. During decoding, we use the beam-search algorithm, setting the beam size to 5. We implement all models in PyTorch \citep{NIPS2019_9015}.
\subsubsection{Baselines}
\textbf{Local and Global Context-Aware Description generator (LoG-CAD)}: proposed by \citet{ishiwatari-etal-2019-learning}, this model achieved the previous state-of-the-art on existing datasets for this task. The model makes use of a BiLSTM \citep{graves2005framewise} to encode sentence-level context, a character-level CNN \citep{zhang2015character} to encode character-level information, and pretrained Google CBOW\footnote{\url{https://code.google.com/archive/p/word2vec/}}\citep{mikolov2013efficient} vectors (for \textsc{Robert} we use the French \texttt{fasttext} word vectors \citep{grave2018learning}). During decoding, this method makes use of a 2-layer attentional 300-dim LSTM decoder with an additional gating mechanism to combine all these sources of encoding information.

\textbf{LSTM baseline (LSTM)}: To show the effect of continous latent variable modeling for this task, and for a more direct comparison to LoG-CAD, we implement an LSTM version of our proposed architecture. Following LoG-CAD, use a 2-layer 300 dimensional BiLSTM as each encoder and use a 10k Byte-Pair tokenized \citep{sennrich-etal-2016-neural} encoder vocabulary. The \textit{neural definition inferer} and the \textit{variational definition modeler} are kept the same as our proposed method. 

\textbf{BERT Baselines}: This baseline is a single-layer attentional 512-dim LSTM-LM decoder conditioned on $\vr_{w_t}$. We use two variants: (1) \textbf{BERT-fr} where $\vr_{w_t}$ is produced by a a frozen BERT-base encoder and (2) \textbf{BERT-ft} where $\vr_{w_t}$ is produced by a BERT-base encoder finetuned during training.

\subsection{Evaluation}

When comparing our approach to our baselines we make use of two automatic evaluation metrics, namely sentence-level BLEU \citep{papineni-etal-2002-bleu, Koehn2007MosesOS} and the recently proposed BERTScore \citep{BERTScore}. While the former is a well-known metric for machine translation, based mainly on n-gram matching between source and target, the latter is a rather new approach that leverages BERT's pretrained contextual embeddings, matching words in candidate and reference sentences by way of cosine similarity. Concretely, BERTScore computes 3 metrics, namely  precision (denoted as $P_{BERT}$), recall (denoted as $R_{BERT}$) and F1 score (denoted as $F_{BERT}$).

Our interest in BERTScore sparks from the fact that it has been recently shown to correlate better with human judgement in system evaluations, and to address the potential issue of coherent definition generations being given low evaluation scores as a result of having zero or low n-gram overlap with the reference sentence. 

Finally, in addition to our automatic evaluation we also performed a human study, where three different human annotators evaluated the output generated by our proposed approach, as well as by the LoG-CAD and BERT-ft baselines. We followed the approach by  \citet{ishiwatari-etal-2019-learning} and used their 1-5 scale: 

\begin{enumerate}
    \item Completely wrong or self-definition
    \item Correct topic with wrong information
    \item Correct but incomplete
    \item Small details missing
    \item Correct
\end{enumerate}
to evaluate 100 randomly sampled instances from \textsc{Oxford}. 

To compare the values obtained for each example across two models, we utilized t-tests and pair-wise bootstrap resampling tests with 10,000 samples \citep{koehnStatisticalSignificanceTests2004}, controlling for the random seed (set to 2 in our experiments).

\section{Results}

\begin{table}
    \scalebox{0.78}{
    \centering
    \begin{tabular}{clllll}
        \toprule
        \textbf{\textsc{Data}} & \textbf{Model} & \textit{BLEU} & $P_{BERT}$ & $R_{BERT}$ & $F_{BERT}$ \\
        \midrule 
	    \multirow{4}{*}{\rot{\textsc{Oxford}}}
            & LoG-CAD & 18.63 & 86.40 & 80.57 & 83.38 \\
          	& LSTM & 21.02 & 85.58 & 85.51 & 85.52\\
            & BERT-fr & 18.26 & 85.95 & 85.11 & 85.50 \\
            & BERT-ft & 27.26 & 87.36 & 87.07 & 87.19 \\
            & VCDM & \textbf{27.38} & \textbf{87.47} & \textbf{87.11} & \textbf{87.27} \\
        \midrule
	    \multirow{5}{*}{\rot{\textsc{Urban}}}
    	    & LoG-CAD & 10.65 & 78.73 & 81.77 & 80.09 \\ 
        	& LSTM & 11.10 & 84.27 & 83.54 & 83.87\\
    	    & BERT-fr & 9.89 & 84.04 & 82.36 & 83.12 \\
    	    & BERT-ft & 11.45 & 84.91 & 82.65 & 83.71 \\ 
	        & VCDM & \textbf{13.90} & \textbf{85.15} & \textbf{83.70}& \textbf{84.36} \\
        \midrule
	\multirow{5}{*}{\rot{\textsc{Wikipedia}}}
            & LoG-CAD & 36.65 & 89.51 & 88.17 & 88.83 \\
            & LSTM & 38.86 & 90.09 & 88.44 & 89.21  \\
            & BERT-fr & 35.97 & 89.51 & 88.11 & 88.77 \\
            & BERT-ft & \textbf{42.97} & 90.48 & \textbf{89.54} & \textbf{89.97} \\
            & VCDM  & 42.27 & \textbf{90.89} & 88.97 & 89.87 \\
        \midrule
	\multirow{5}{*}{\rot{\textsc{Cambridge}}}
            & LoG-CAD & 16.87 & 86.09 & 85.32 & 85.68\\ 
        	& LSTM & 16.44 & 86.21 & 85.43 & 85.81\\
            & BERT-fr & 17.90 & 87.17 & 85.95 & 86.53 \\
            & BERT-ft & 20.04 & 87.81 & 86.88 & 87.24 \\
            & VCDM & \textbf{22.46} & \textbf{88.16} & \textbf{87.46} & \textbf{87.70} \\
        \midrule
	\multirow{5}{*}{\rot{\textsc{Robert}}}
        & LoG-CAD & 22.94 & 69.77 & 68.09 & 68.80\\ 
    	& LSTM & 39.76 & 78.89 & 79.18&78.90 \\
        & BERT-fr &23.61&73.74&71.90& 72.63\\
        & BERT-ft & 41.50 & 81.82 & 80.54 & 81.02\\
        & VCDM & \textbf{44.97} & \textbf{82.80} & \textbf{81.96} & \textbf{82.24}\\
    \bottomrule
    \end{tabular}
    }
    \caption{Results on the test set for \textsc{Urban, Oxford, Wikipedia, Cambridge}, and \textsc{Robert}.} \label{tb:results}
\end{table}

\textbf{Automatic Evaluation}: Table~\ref{tb:results} shows the results on the test set for each reported metric and dataset. Firstly, we note that the LSTM Baseline is able to consistently outperform LoG-CAD in terms of BERTScore, although with mixed results in terms of BLEU. We think this difference is mainly due to the n-gram matching nature of BLEU, which tends to give better scores for longer but incorrect generations, as the example in Table \ref{tb:bert-score} shows, while also being unable to adequately handle cases where the definitions are expressed using words not present in the gold standard. We believe these results validate the usage of a metric such as BERTScore on this task, ultimately showing that tackling definition modeling with a generative approach can lead to improved results, and suggesting that the incorporation of a latent variable that models the underlying definition space is beneficial for this task. 

\begin{table}
    \centering
    \footnotesize
    \scalebox{1.0}{
    \setlength{\tabcolsep}{2pt}
    \begin{tabular}{l L{0.35\textwidth} }
        \toprule
        Word      & Frankenstein \\
        Context   & In arming the dictator, the US was creating a \textit{\textbf{Frankenstein}} \\
        Reference & something that destroys or harms the person or people who created it \\
    \end{tabular}}
    \setlength{\tabcolsep}{3pt}
    \scalebox{0.9}{
    \begin{tabular}{L{0.25\textwidth} c c c c}
        \toprule
        Generated & BL & P & R & F \\
        \midrule
        something that you say or do that you think someone of something is ridiculous & \textbf{12.5} & 83.41 & 84.78 & 84.09 \\
        \midrule
        an extremely frightening or offensive person & 8.13 & \textbf{87.00} & 84.78 & \textbf{85.88} \\
        \bottomrule
    \end{tabular}}
     \caption{An example from the \textsc{Cambridge} test set, showing an evaluation issue caused by BLEU. The generated outputs of the LoG-CAD baseline are shown above, and ours below. BL stands for sentence BLEU and P, R and F stand for $P_{BERT}$, $R_{BERT}$, and $F_{BERT}$.}
     \label{tb:bert-score}
\end{table}
\begin{table}
    \footnotesize
    \centering
	\begin{tabular}{lllll}
        \toprule
        \textbf{Configuration} &
        $\mathbf{F_{BERT}}$ ($\Delta$) & \textbf{BLEU} ($\Delta$) \\ \midrule
        \textbf{VCDM} & \textbf{87.70} ($\;\,$---$\;\,\,$) & \textbf{22.46} ($\;\,$---$\;\,\,$) \\
        Decoder LSTM Cell & 87.68 (-0.02) & 22.27 (-0.19) \\
        Frozen definition encoder & 87.14 (-0.56) & 20.70 (-1.76) \\
        Tied encoders & 87.14 (-0.56) & 20.59 (-1.87) \\
        Frozen encoders & 86.35 (-1.35) & 17.42 (-5.04) \\
        Frozen context encoder & 85.19 (-2.51) & 13.49 (-8.97) \\
        \bottomrule
        \end{tabular}
    \caption{Results of the ablation study performed on \textsc{Cambridge}.}
    \label{tb:ablation}
    \end{table}

Results on Table \ref{tb:results} also show that the inclusion of BERT significantly improves generation quality in terms of BERTScore and BLEU on most datasets. This suggests that the inclusion of pretrained deep contextual word representations is beneficial for the task, which is expected given its contextual nature. We also see that VCDM is able to successfully leverage BERT, as our model is able to offer improved results compared to BERT baselines in all datasets except \textsc{Wikipedia}. We think these results offer additional empirical evidence to support the effectiveness of our generative approach. Improvements provided by our model are particularly significant in the case of \textsc{Urban}, a dataset which there are many rare words and the context is arguably less informative due to its noisy properties.


We surmise that the subpar performance of VCDM over BERT-ft in \textsc{Wikipedia} is related to the properties of the dataset domain (i.e. description generation of named entities). With this in mind, it could be argued that a completely context- focused architecture (such as that of our finetuned BERT baseline) has properties that are more beneficial in this setting. Contrary to findings in \citet{ishiwatari-etal-2019-learning} which argue for the inclusion of a global context during generation, we find that a contextually-focused (local context) architecture with a strong context encoder (such as BERT) results in better performance within this domain.

\textbf{Ablation Study}: To further evaluate the contribution of each introduced component in our approach we performed an ablation study on \textsc{Cambridge}. Results of these experiments are summarized in Table~\ref{tb:ablation}, where it is possible to see that each of our introduced components is beneficial to the task. Note that the VCDM-Cell vs LSTM-Cell improvement is minimal on this dataset. The purpose of the integration of the latent variable in the decoder LSTM cell is for it to act as a "global definition signal" so we can rely on the properties of the latent variable. As this property is especially useful in cases in which there is noisy context, we think it is reasonable to assume that as context here is more informative, the performance gain from including a global definition signal is relatively small. We also see that freezing the context encoder has a extremley negative impact on performance. We believe this is because the context-encoder hasn't effectively learned to use the \textit{phrase-context pairs}~(Sec.~\ref{sec:pcp}).

\begin{table}
    \centering
    \footnotesize
    \begin{tabular}{llll}
        \toprule
        Model   & Model  & p-value bootstrap & p-value t-test \\
        \midrule
        VCDM    & LoG-CAD & 0.005 & $1.4 \times 10^{-8}$\\
        BERT-ft & LoG-CAD & 0.006 & $4.2 \times 10^{-5}$ \\
        VCDM    & BERT-ft & 0.797 & $1.6 \times 10^{-2}$ \\
        \bottomrule
    \end{tabular}
    \caption{Exact p-values of the performed statistical tests, to compare the scores obtained during our human evaluation.}
    \label{table:p-values}
\end{table}
\textbf{Human Evaluation}: Average human scores obtained are $2.51$ for LoG-CAD, $3.08$ for BERT-ft and \textbf{$\mathbf{3.31}$ for VCDM}. When tested for statistical significance (Table~\ref{table:p-values}), we observed that both VCDM and BERT-ft were superior to LoG-CAD with 99\% confidence, using both paired t-tests or pair-wise bootstrap resampling tests \citep{koehnStatisticalSignificanceTests2004}, and that the difference between VCDM and BERT-ft was statistically significant at 95\% for the t-test. 
\begin{table}
    \tabcolsep=0.11cm
    \footnotesize
    \begin{tabular}{cp{0.4\linewidth}p{0.34\linewidth}}
    \toprule[0.15em]
         \textbf{Word:} & Present(VB) & Present(NN)\\
         \midrule[0.05em]
         \textbf{Context:}& Within a sexist ideology and a male-dominated cinema, the woman is \textbf{\textit{presented}} as what she represents for man. &  In addition to this, think of the \textbf{\textit{presents}}, the toys, gift sets, and most importantly, all that wrapping paper. \\
         \midrule[0.05em]
         \textbf{Reference:}& To represent (someone or something) to others in a particular way & A thing given to someone as a gift \\
         \midrule[0.1em]
         \textbf{LoG-CAD:} & a person who is present in a particular way & a person's mind\\ \midrule[0.05em]
         \textbf{BERT-ft:} & Portray or regard (someone) as a particular person, idea or action&An item of furniture presented to resemble a bride $<$unk$>$ \\ \midrule [0.05em]
         \textbf{VCDM:} & To portray or describe (someone or something) in a particular context & A thing kept as a gift for children \\
         \bottomrule[0.15em]
    \end{tabular}
    \caption{Example showing the generated definitions for two senses of the word ``present'', taken from \textsc{Oxford}.}
    \label{tb:example}
\end{table}

\textbf{Qualitative Evaluation}: Finally we provide a qualitative evaluation by showing an example of the output of our model and of two of our baselines, in Table~\ref{tb:example}. In the \textsc{Oxford} dataset, which this example is taken from there are 7 senses of the ``present", showing VCDM's ability to effectively disambiguate between a large amount of senses.

\section{Conclusion}
In this paper we have introduced a generative model that directly combines distributional and lexical semantics via a continuous latent variable for the task of definition modeling. Empirical results on multiple corpora, including two new datasets released, show that our model is able to outperform previous work by a consistent margin, also successfully being able to leveraging contextualized word representations. For future work we are interested in exploring how definition modeling could be adapted to a multilingual or cross-lingual setting.



\section*{Acknowledgments}
We are grateful for the support provided by the NVIDIA Corporation, donating two of the GPUs used for this research. We thank Victor Zhong for insightful discussions, and thank Pablo Loyola, Cristian Rodriguez-Opazo, and Jorge Balazs for proofreading the work and providing useful suggestions.

\bibliography{emnlp2020}
\bibliographystyle{acl_natbib}

\appendix
\end{document}


\maketitle

\appendix


\section{Datasets}

Tables \ref{tb:data_oxford}, \ref{tb:data_urban} and \ref{tb:data_wikipedia} provide a summary of the sizes of each partition for the datasets \textsc{Oxford}, \textsc{Urban} and \textsc{Wikipedia}, respectively.

\begin{table}[h!]
	\centering
    \scalebox{0.65}{
    \begin{tabular}{clllll}
        \toprule
        \multirow{2}{*}{\textbf{Partition}} & \multicolumn{2}{c}{\textbf{Count}} & \multicolumn{3}{c}{\textbf{Length}} \\
        \cmidrule{2-6}
        & \textbf{Phrases}     & \textbf{Examples}     & \textbf{Phrase} & \textbf{Context} & \textbf{Definition} \\
        \midrule
    	Train & 33,128 & 97,855  & 1.00 & 17.74 & 11.02  \\
    	Valid & 8,867 & 12,232 & 1.00 & 17.80 & 10.99 \\
    	Test & 8,850 & 12,232 & 1.00 &  17.56 & 10.95 \\
    	\bottomrule
        \end{tabular}
    }
    \caption{Statistics for \textsc{Oxford}. The number of individual phrases, number of examples, and the mean lengths of each partition of the dataset are reported.}
    \label{tb:data_oxford}
\end{table}

\begin{table}[h!]
	\centering
    \scalebox{0.65}{
    \begin{tabular}{clllll}
        \toprule
        \multirow{2}{*}{\textbf{Partition}} & \multicolumn{2}{c}{\textbf{Count}} & \multicolumn{3}{c}{\textbf{Length}} \\
        \cmidrule{2-6}
        & \textbf{Phrases}     & \textbf{Examples}     & \textbf{Phrase} & \textbf{Context} & \textbf{Definition} \\
        \midrule
        Train  & 190,696 & 411,384 & 1.54 & 10.89 & 10.99 \\
        Valid  & 26,876 & 57,883 & 1.54 & 10.86 & 10.95 \\
        Test   & 26,875 & 38,371 & 1.68 & 11.14 & 11.50 \\
    	\bottomrule
        \end{tabular}
    }
    \caption{Statistics for \textsc{Urban}. The number of individual phrases, number of examples, and the mean lengths of each partition of the dataset are reported.}
    \label{tb:data_urban}
\end{table}

\begin{table}[h!]
	\centering
    \scalebox{0.65}{
    \begin{tabular}{clllll}
        \toprule
        \multirow{2}{*}{\textbf{Partition}} & \multicolumn{2}{c}{\textbf{Count}} & \multicolumn{3}{c}{\textbf{Length}} \\
        \cmidrule{2-6}
        & \textbf{Phrases}     & \textbf{Examples}     & \textbf{Phrase} & \textbf{Context} & \textbf{Definition} \\
        \midrule
        Train  & 151,995 & 887,455 & 2.10 & 18.79 & 5.89 \\
        Valid  & 8,361 & 44,003 & 2.11 & 19.21 & 6.31 \\
        Test   & 8,397  & 57,232 & 2.10 & 19.02 & 6.94 \\
    	\bottomrule
        \end{tabular}
    }
    \caption{Statistics for \textsc{Wikipedia}. The number of individual phrases, number of examples, and the mean lengths of each partition of the dataset are reported.}
    \label{tb:data_wikipedia}
\end{table}

\section{Evaluation}

We make use of the \texttt{sentence-bleu.cpp}\footnote{\url{https://github.com/moses-smt/mosesdecoder/blob/master/mert/sentence-bleu.cpp}} script in the MOSES \citep{Koehn2007MosesOS} GitHub repository to compute sentence-level BLEU, and use the \texttt{bert-score} Python package\footnote{\url{https://github.com/Tiiiger/bert_score}} to calculate BERTScore, with hash \small{\texttt{roberta-large\_L17\_no-idf\_version=0.3.2 (hug\_trans=2.8.0)}} \normalsize for the English datasets and hash \small{\texttt{bert-base-multilingual-cased\_L9\_no-idf\_ version=0.3.3(hug\_trans=2.10.0)}} \normalsize for \textsc{Robert} which is in French.
\normalsize
For datasets where there are multiple examples for a given word sense, such as \textsc{Wikipedia}, we note that results provided by \cite{ishiwatari-etal-2019-learning} are obtained by first averaging the evaluation metrics for multiple examples for a given sense ---although this is not reported on the paper--- which tends to inflate the final values of the metrics. Instead, in this paper report example-wise metric averages, which provide more realistic values of the aggregated evaluation metrics. 



\section{Model Details}
We train each of our models using a batch size of 64, and set 1e-3 as the initial learning rate for the Adam optimizer. However, when finetuning BERT (or CamemBERT) in any circumstance, we set the initial learning rate for the BERT parameters to be 5e-5 and use a linear warmup schedule, warming up for the first epoch. We train all models in PyTorch \citep{NIPS2019_9015}, and use the HuggingFace\footnote{\url{https://github.com/huggingface/transformers}} \citep{wolf2019huggingfaces} implementation of CamemBERT-base and BERT-base-uncased. Trainable parameter counts are for each model are shown in Table~\ref{tb:param_count}.

Additionally, we re-implement LoG-CAD \citep{ishiwatari-etal-2019-learning} using the authors' GitHub repository\footnote{\url{https://github.com/shonosuke/ishiwatari-naacl2019}}.

\section{Infrastructure and Environment}

Experiments for different datasets were run in two different machines:

\begin{itemize}
    \item A server machine with an Intel Xeon E5-2630 CPU, and two NVIDIA RTX-2080 (Driver 418.56, CUDA 10.1)  GPUs, running Ubuntu 16.04

    \item An additional server machine with an Intel Core i7-6850K CPU and two NVIDIA Titan Xp (Driver 430.50, CUDA 10.1) GPUs, also running Ubuntu 16.04 

\end{itemize}


\section{Additonal Output Examples}
\begin{table}[h!]
    \tabcolsep=0.11cm
    \footnotesize
    \begin{tabular}{cp{0.8\linewidth}}
    \toprule[0.15em]
         \textbf{Word:}& Yen\\
         \midrule[0.05em]
         \textbf{Context:}& If Koizumi has enjoyed some economic success, say critics, it has been through a combination of good luck and what many believe has been an artificial weakens of they yen against the dollar.\\
         \midrule[0.05em]
         \textbf{Reference:}&the basic monetary unit of japan. \\
         \midrule[0.1em]
         \textbf{LoG-CAD:} & a longing or yearning\\ \midrule[0.05em]
         \textbf{BERT-ft:} & a monetary in a foreign country \\ \midrule [0.05em]
         \textbf{VCDM:} & the basic monetary unit of japan, equal to 100 cents.\\
         \bottomrule[0.15em]
    \end{tabular}
    \caption{Descriptions for the rare word ``yen"}    \label{tb:example}
\end{table}
\begin{table}[h!]
    \tabcolsep=0.11cm
    \footnotesize
    \begin{tabular}{cp{0.8\linewidth}}
    \toprule[0.15em]
         \textbf{Word:}& doucheturd\\
         \midrule[0.05em]
         \textbf{Context:}& Brad, you're such a doucheturd.\\
         \midrule[0.05em]
         \textbf{Reference:}&insulting noun, being both a douche or douchebag and a turd \\
         \midrule[0.1em]
         \textbf{LoG-CAD:} &  a person who is a douchebag\\ \midrule[0.05em]
         \textbf{BERT-ft:} &  a person who is a douchebag \\ \midrule [0.05em]
         \textbf{VCDM:} &  a person who is a mix of a douche and a turd\\
         \bottomrule[0.15em]
    \end{tabular}
    \caption{Descriptions for the slang word ``doucheturd"}    \label{tb:example}
\end{table}

\begin{table*}[h!]
    \scriptsize
    \centering
    \begin{tabular}{r L{0.6\textwidth}}
        \toprule
        Gold  & ( especially of a political party ) sponsor ( a candidate ) in an election \\
        LoG-CAD & a candidate or candidate in a race or election \\
        VCDM       & set ( a person or team ) in an election \\
        \midrule
        Gold  & ( of a batsman ) run from one wicket to the other in scoring or attempting to score a run . \\
        LoG-CAD & a race or contest in which a race is run \\
        VCDM       & ( of a sports team ) run by hitting at the reach of the three runs \\
        \midrule
        Gold  & a large open stretch of land used for pasture or the raising of stock\\
        LoG-CAD & ( of a horse ) be $<$\text{unk}$>$\\
        VCDM       & a specially $<$\text{unk}$>$ area for cattle or cattle\\
        \midrule
        Gold  & a large open stretch of land used for pasture or the raising of stock\\
        LoG-CAD & ( of a horse ) be $<$\text{unk}$>$\\
        VCDM       & a specially $<$\text{unk}$>$ area for cattle or cattle\\
        \midrule
        Gold  & a preliminary test of a procedure or system \\
        LoG-CAD & a person or thing that is $<$\text{unk}$>$ or $<$\text{unk}$>$ \\
        VCDM       & a continuous search or undertaking \\
        \midrule
        Gold  & a race between candidates for elective office\\
        LoG-CAD & a series of people who are $<$\text{unk}$>$\\
        VCDM       & be the charge of \\
        \midrule
        Gold  & a row of unravelled stitches \\
        LoG-CAD & a short , narrow $<$\text{unk}$>$ \\
        VCDM       & a $<$\text{unk}$>$ in a careless or $<$\text{unk}$>$ way \\
        \midrule
        Gold  & a track made or regularly used by a particular animal \\
        LoG-CAD & move or cause to move in a specified direction \\
        VCDM       & a $<$\text{unk}$>$ or $<$\text{unk}$>$ run over a tree \\
        \midrule
        Gold  & become undone \\
        LoG-CAD & be $<$\text{unk}$>$\\
        VCDM       & become undone by being undone\\
        \midrule
        Gold  & cause something to pass or lead somewhere\\
        LoG-CAD & move or move in a $<$\text{unk}$>$\\
        VCDM       & cause something to pass or lead somewhere by constant strength\\
        \midrule
        Gold  & cause to perform\\
        LoG-CAD & make a series of facts or plans\\
        VCDM       & put in an area \\
        \midrule
        Gold  & diarrhoea .\\
        LoG-CAD & a person 's $<$\text{unk}$>$ \\
        VCDM       & a $<$\text{unk}$>$ of $<$\text{unk}$>$ \\
        \midrule
        Gold  & emit or exude a liquid \\
        LoG-CAD & ( of a person 's eyes ) move or cause to move in a specified direction \\
        VCDM       & ( of a person ' s $<$\text{unk}$>$ ) become $<$\text{unk}$>$ with $<$\text{unk}$>$\\
        \midrule
        Gold  & extend or continue for a certain period of time \\
        LoG-CAD & be a result of \\
        VCDM       & pass for a certain time of time \\
        \midrule
       Gold  & fail to stop at ( a red traffic light ) \\
        LoG-CAD & ( of a vehicle ) move or move in a specified direction \\
        VCDM       & run at or at a particular $<$\text{unk}$>$ \\
        \midrule
        Gold  & move about in a hurried and hectic way \\
        LoG-CAD & be in a specified way \\
        VCDM       & go or run somewhere in a particular place \\
        \midrule
        Gold  & move or cause to move between the spools of a recording machine \\
        LoG-CAD & move or move in a specified direction \\
        VCDM       & use or enable ( a container ) to a desired point in a specified direction \\
        \midrule
        Gold  & publish or be published in a newspaper or magazine \\
        LoG-CAD & a $<$\text{unk}$>$ or $<$\text{unk}$>$ \\
        VCDM       & ( of a newspaper or a public station ) publish or broadcast ( a television programme )      \\
        \midrule
        Gold  & put ( a form of public transport ) in service \\
        LoG-CAD & ( of a vehicle ) be $<$\text{unk}$>$ or $<$\text{unk}$>$ \\
        VCDM       & provide ( an undertaking , train , or service ) for a service \\
        \midrule
        Gold  & the act of running ; traveling on foot at a fast pace \\
        LoG-CAD & a run in a race \\
        VCDM       & the act of running ; traveling on foot at a fast pace \\
        \midrule
        Gold  & the act of testing something \\
        LoG-CAD & the act of $<$\text{unk}$>$ something \\
        VCDM       & the act of testing something \\
        \midrule
        Gold  & travel a route regularly \\
        LoG-CAD & move or move in a specified direction \\
        VCDM       & travel a route regularly \\
        \midrule
        Gold  & the after part of a ship 's bottom where it rises and narrows towards the stern .\\
        LoG-CAD & a $<$\text{unk}$>$ or $<$\text{unk}$>$ . \\
        VCDM       & an act of $<$\text{unk}$>$ a boat ' s foot \\
        \midrule
        Gold  & the average or usual type of person or thing \\
        LoG-CAD & a person 's existence or belief \\
        VCDM       & the general aspects of something , especially a language \\
        \bottomrule
    \end{tabular}
    \caption{Comparison of the outputs for the 23 senses of ``run' on \textsc{Oxford}.}
\end{table*}

\clearpage
\newpage

\bibliography{emnlp2020}
\bibliographystyle{acl_natbib}